\documentclass[runningheads]{llncs}
\usepackage{graphicx}
\usepackage{algorithm}
\usepackage{algorithmic}
\usepackage{amsmath}
\usepackage{amsfonts}
\usepackage{comment}
\usepackage{subfigure}
\usepackage{bm}

\newcommand{\E}{\mathbb{E}}

\DeclareMathOperator*{\argmin}{argmin}
\DeclareMathOperator*{\argmax}{argmax}

\begin{document}
\title{Action Set Based Policy Optimization for \\ Safe Power Grid Management}
%
%
\author{\textbf{Bo Zhou*, Hongsheng Zeng*, Yuecheng Liu, Kejiao Li, Fan Wang, Hao Tian}}
\authorrunning{\textit{Zhou, Zeng et al.}}
%
\institute{Baidu Inc., China \\
\{zhoubo01, zenghongsheng, liuyuecheng, likejiao, wang.fan, tianhao\}@baidu.com}
\maketitle              
\begin{abstract} 
\let\thefootnote\relax\footnotetext{*  Equal Contribution.}
Maintaining the stability of the modern power grid is becoming increasingly difficult due to fluctuating power consumption, unstable power supply coming from renewable energies, and unpredictable accidents such as man-made and natural disasters. As the operation on the power grid must consider its impact on future stability, reinforcement learning (RL) has been employed to provide sequential decision-making in power grid management. However, existing methods have not considered the environmental constraints. As a result, the learned policy has risk of selecting actions that violate the constraints in emergencies, which will escalate the issue of overloaded power lines and lead to large-scale blackouts. In this work, we propose a novel method for this problem, which builds on top of the search-based planning algorithm.
At the planning stage, the search space is limited to the action set produced by the policy. The selected action strictly follows the constraints by testing its outcome with the simulation function provided by the system. At the learning stage, to address the problem that gradients cannot be propagated to the policy, we introduce Evolutionary Strategies (ES) with black-box policy optimization to improve the policy directly, maximizing the returns of the long run. In NeurIPS 2020 Learning to Run Power Network (L2RPN) competition, our solution safely managed the power grid and ranked first in both tracks. 

\keywords{Power Grid Management  \and Reinforcement Learning \and Planning.}
\end{abstract}
\section{Introduction}
Electrical grid plays a central role in modern society, supplying electricity across cities or even countries. However, managing the well functioning of the power network not only suffer from the fluctuating power consumption and unexpected accidents in the network, but also faces challenges from the unprecedented utilization of renewable energy \cite{yang2019two,marot2020learning}. The control system has lower operational flexibility as more renewable energy power plant connects to the power grid. For example, wind power stations that rely on seasonal wind cannot provide such stable electricity throughout the year like traditional thermal power station. Other issues such as rapidly growing electric car deployment that increases fluctuations in electricity consumption across regions also pose new challenges.

There are many efforts on applying deep reinforcement learning (RL) in power grid management, the recent technique emerged as a powerful approach for sequential decision-making tasks \cite{mnih2015human,schrittwieser2020mastering,lee2020learning}. Taking the grid states as input, the policy adjusts the power generation of each power plant to feed the loads safely  \cite{ernst2004power,diao2019autonomous,yang2019two}. To further improve operational flexibility, recent works also study on managing the power grid through topological actions (e.g., reconfiguring bus assignments and disconnecting power lines) \cite{fisher2008optimal,khodaei2010transmission,marot2020learning}.

While RL-based approaches for grid management have achieved impressive results, existing methods ignore the environmental constraints. In power grid management, there are a number of complex rules that the selected actions must follow. For example, the system should avoid operations that lead to the disconnection of some residents to the power grid. Based on such constraints, an intuitive solution is to discourage actions that violate the rules by adding penalties on the feedback. However, this approach does not guarantee that all the actions produced by the policy strictly satisfy the constraints \cite{shah2020solving}.

In this work, we propose an action set based method to manage the power grid safely through topological operations while strictly meeting the environmental constraints. The algorithm builds on top of the search-based planning approach, which has recently shown performance that exceeds human in challenging tasks \cite{silver2016mastering,schrittwieser2020mastering}. At the planning stage, rather than use the entire action space as the search space, we limit the search space to the action set produced by the policy model. We then test the outcome of each candidate action in the action set with the simulation function provided by the system and select the action that strictly meet the constraints. However, such module blocks the gradient route of back-propagation from the selected action to the policy. To address the problem, we introduce Evolutionary Strategies (ES) \cite{salimans2017evolution} to directly optimize the policy towards maximizing long-term returns, by regarding the planning stage as block-box. Our agent participated in NeurIPS 2020 L2RPN competition, which provides two challenging power grid environments: Track 1 with unexpected power line attacks and Track 2 with increasing proportion of renewable energy plants. We ranked 1st place in both tracks.\footnote{Our code is available open-source at:  https://github.com/PaddlePaddle/PARL\\/tree/develop/examples/NeurIPS2020-Learning-to-Run-a-Power-Network-Challenge} \\

\section {Related Work}
Traditional system protection scheme (SPS) \cite{bernard1996735,trudel1999hydro,otomega2007distributed,tomsovic2005designing} builds an expert system to maintain the electrical flow and voltage magnitudes within safe range. The system relies on the network state such as voltage and electricity load level to make the decision. If overloaded power lines are detected, the system is triggered to take actions following the expert rules. SPS has less computation complexity and can provide real-time operations. The limitation of SPS is that not all possible issues can be foreseen at the stage of designing the system, which may result in instability and eventually lead to large-scale collapse \cite{larsson2002emergency}.

A number of approaches formulate the power network management as a control problem and apply control and optimization theory to solve it. \cite{jin2009model} uses model predictive control (MPC) \cite{garcia1989model} to select actions by minimizing the cost of operations and voltage-deviations under the security constraints of the power network, with a linear approximation model for state prediction. \cite{larsson2002emergency} predicts the future states based on a simulation model with nonlinear differential-algebraic equations and adopts tree search for optimization. However, The performance of MPC-based often heavily relies on the accuracy of the dynamics model \cite{huang2019adaptive}. 

Prior work has also modeled the grid management as a Markov decision process (MDP) and adopted reinforcement learning for sequential decision making. \cite{ernst2004power,huang2019adaptive,diao2019autonomous} operates the power network by adjusting generator outputs or reducing the load of electricity. \cite{dalal2016hierarchical} proposes a hierarchical architecture to consider long-term reliability and provide real-time decision making, where the policy updates at a fast time-scale and the value function updates at a slow time-scale. To further improve the operation flexibility, recent research \cite{fisher2008optimal,khodaei2010transmission,marot2020learning} studies on reconfiguring the topology for power grid management (e.g., switching the bus assignments of the loads and generators in a substation). \cite{yoon2021winning} employs the after-state representation \cite{sutton2018reinforcement} to reduce the difficulty of modeling the large observation and action space, with a hierarchical policy to determine and adjust the network topology. 

\begin{figure}[h]
  \centering
  \includegraphics[width=\linewidth]{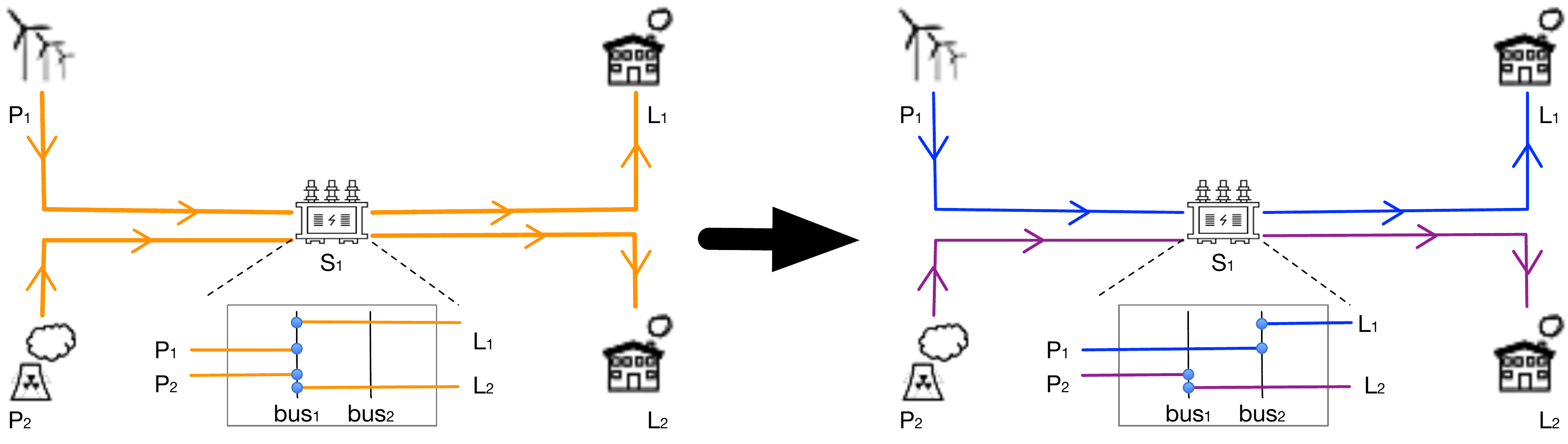}
  \caption{Illustration of topology reconfiguration through the two buses in the substation. (Left) The two power stations provide electricity to two loads simultaneously, with the solid blue dots representing the connection between the object and the bus. (Right) Each power plant only transmits power to one load after the reconfiguration.}
  \label{fig:topo_act}
\end{figure}

\section{Preliminary}
In this section, we first formulate the objective of power grid management. Then we consider the task as a Markov decision process (MDP) and introduce the basic idea of search-based planning algorithm \cite{coulom2006efficient,silver2016mastering} proposed for solving sequential decision tasks. Our algorithm builds on top of the search-based planning approach.

\subsection{Power Grid Management}
There are three basic and important elements in a power grid: power plants $\bm{P}$ (indexed from $P_1...P_N$), substations $\bm{S}$ (indexed from $S_1...S_M$) and loads $\bm{L}$ (indexed from $L_1...L_K$). All the elements are connected together with the power lines that transmit power from one end to the other. There are two bus bars in each substation, and every element connected to the substations must be connected to one of them. We provide a toy example in Figure~\ref{fig:topo_act} to demonstrate how to reconfigure the topology by switching the connected bus bar of the elements. There is a small grid with two power stations, where two stations provide power for two loads through the substation, and all the elements are connected to bus 1. If we reconnect $P_1$ and $L_1$ to the second bus, then each plant only provides power for only a load. \\
One of the most distinguished features of grid management is the topological graph that describes how electrical elements construct the power network. The grid can be represented as an undirected graph $G (V, E)$, where $V = (\bm{P}, \bm{S}, \bm{L})$ is the node-set composed of all the basic elements, and $E$ is the edge set representing the connections of elements. Each edge $e_i(u, v, t_u, t_v)$ in $E$ represents that a power line connects $u\in V$ with $v \in V$, and $t_u,t_v\in \{0,1,2\}$ indicate the bus to which the power line connects. 0 means that the line is not connecting with a substation, while 1 and 2 represent that the power line connects to the first and second bus, respectively. Denote the number of elements connected to each substation $S_i$ by $|S_i|, i=1,2,..,M$. The number of possible topologies generated by bus switching in a substation is $2^{|S_i|}$, and for the whole power grid, there are $2^{\sum_{i=1}^{i=M}|S_i|}$ possible topology. \\
The task of operating the power grid is to maintain electrical flows of power lines within the acceptable range. If the current exceeds the maximum limit and maintains a high level, it will damage the power line and increase the transmission burden of other power lines, which may cause a large-scale power outage. For each power line $e_i \in E$, the ratio of current flow over maximum flow should stay at a safe level less than 1. The controller/operator can take three types of actions to avoid or address the overflow issue: (1) reconfiguring the bus connection in the substations (2) disconnecting or connecting the power lines (3) adjusting the power generation of the plants. \\
There are mainly two metrics for evaluating the controllers: the time horizon that power grid runs safely and the cost of operations decided by the controller (e.g., reconfiguring the grid topology and adjusting the power generation has different cost). The total reward of a controller can be defined as: $R = \sum_{t=0}^T r_t - c_t$, where $r_t > 0$ is the positive feedback and $c_t$ is the operation cost. If any load or generator is disconnected from the power grid, the controller will not receive any reward since it takes dangerous actions that collapse the grid.\\
To encourage the study in power grid management, Coreso (European RSC) and RTE (French TSO) built a simulation environment named Grid2Op that runs the power system with real-time operations \cite{marot2020learning}. Grid2Op provides a training data set of multiple hundreds of yearly scenarios at 5 minutes resolution. More details about the environment can be found in Appendix~\ref{sec:env}.

\subsection{Search-based Planning}
We now consider power grid management as a MDP, defined by the tuple $(\mathcal{S}, \mathcal{A}, P, r, \gamma, \rho_0)$. $\mathcal{S}$ and $\mathcal{A}$ represent the state and action spaces, respectively. Note that $\mathcal{S}$ includes not only the topology graph mentioned in Section~3.1 but also other grid information such as power generation of the plants. We denote the distribution of initial states as $\rho_0$, the environment dynamics as $P(s_{t+1}|s_t, a_t)$. The reward function $r(s_t, a_t)$ relies on the state and action, and the discount factor $\gamma \in (0, 1)$ is used for accumulative reward computation. \\
At each time step t, the controller selects an action $a_t \in \mathcal{A}$ following the policy $\pi$, then the environment transits into the next state according to P. The optimal policy to obtain the maximum average reward can be defined:
\begin{align}
    \pi^*=\argmax_{\pi} \E_{a_t\sim \pi,s_0\sim\rho_0,s_{t+1}\sim P} \sum_{t=0}^\infty \gamma^t r(s_t, a_t)
    \label{eq:rl_object}
\end{align}
Note that the selected actions must meet the power network constraints.

To obtain the optimal policy, we define the state-action value function $Q(s, a)$ to estimate the discounted future reward at state $s$ after taking action $a$.
Once the optimal $Q*(s,a)$ is learned, the optimal policy can be obtained by taking the action with the maximum estimated Q value at each step: $a^* = \argmax_{a \in \mathcal{A}} Q^*(s,a)$.

To learn the optimal Q function, we can adopt Monte Carlo sampling methods \cite{bishop2006pattern} and bootstrapping approaches \cite{sutton2018reinforcement} such as Q-learning \cite{watkins1992q}. The search-based planning algorithm combines the Monte Carlo method with tree search to gradually approximate the optimal Q function. At the search stage, the tree is traversed by simulation, starting from the root node of state $s_t$. At each simulation step, an action $a_t$ is sampled from the search policy until a leaf node $S_L$ is reached:
\begin{align}
\label{eq:planning_act_selection}
a_t = \argmax_a(Q(s_t, a) + U(s_t, a))
\end{align}
where $U(s_t, a)  \propto \frac{1}{N(s_t, a)}$ is a bonus that decays with increasing visit count $N(s_t, a)$ to encourage exploration \cite{silver2016mastering}. The bonus term can also be combined with a prior policy that matches the search policy: $U(s, a) \propto \frac{P(s,a)}{N(s_t, a)}$ \cite{silver2016mastering,schrittwieser2020mastering}. The future return of the leaf node is evaluated by a value function $V(s)$.

At the learning stage, The state action function $Q(s, a)$ and visit count $N(s, a)$ are updated along the traversed nodes. For the simulation trajectory $\tau(s_0, a_0, r_0,...,s_{l-1}, a_{l-1}, r_{l-1})$ with length $l$, we can estimate the discounted future reward with the value function $V(s)$: $G(s_0, a_0) = \sum_{t=0}^{t=l-1}r_0 \gamma ^t r_t + \gamma^l V(s_{l-1})$ \cite{schrittwieser2020mastering}. For each edge $(s, a)$ in the simulation path, we can perform the following update:
\begin{align}
Q(s, a) &= \frac{N(s,a) * Q(s, a) + G(s, a)}{N(s,a)+1} \\
N(s, a) &= N(s, a) + 1.
\end{align}
The value function $V(s)$ is updated through supervised learning that fits the average return starting from the state $s$.

\section {Methodology}
\begin{figure*}[t]
    \centering
    \includegraphics[width=\linewidth]{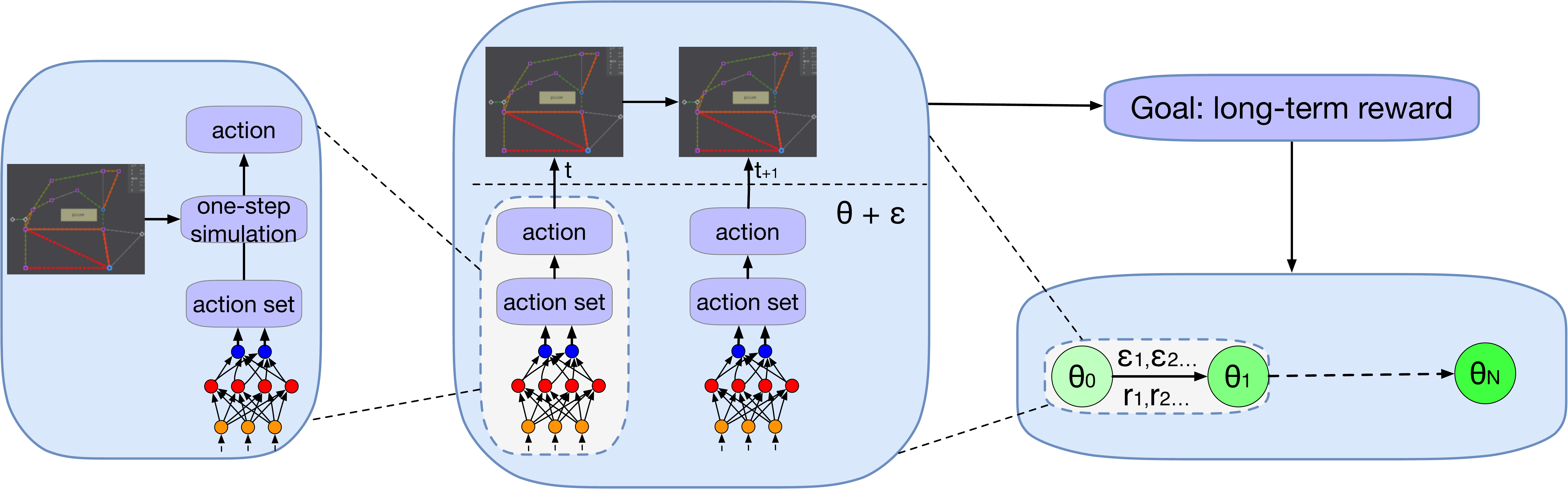}
    \caption{Searching with the action set produced by the policy. (Left) At the search stage, the simulation rolls out the trajectory by searching the action candidates produced by the policy. (Middle) An exploration policy with parameter noise interacts with the environment and collects the feedback from the environment. (Right) The policy is updated towards maximizing the average future reward over the exploration and search trajectories.}
    \label{fig:overview}
\end{figure*}

We now introduce a novel search-based planning algorithm that performs Search with the Action Set (SAS). The goal of the algorithm is to maximize the average long-term reward. We will detail how to optimize the policy towards this goal while strictly meeting the environment constraints. The overview of SAS is summarized in Figure~\ref{fig:overview}. 
\subsection{Search with the Action Set}
At each simulation step, the policy $\pi_{\theta}(a_t|s_t)$ parameterized by $\theta$ outputs a vector of probabilities that the actions should be selected, and the top K actions with higher probabilities form the action-set $A$. We then leverage the simulation function $f_s(s_t, a)$ for action selection to ensure that the action strictly meets the constraints and rules, by simulating the outcome of each action $a \in A$ and filtering actions that violate the constraints. For notational simplicity, we denote the filtered action set by $A$ again. Finally, the algorithm selects an action from $A$ based on the value function $V(s_t)$:
\begin{align}
a_t = \argmax_{a \in A}(V(f_s(s_t,a)).
\label{eq:action_selection_0}
\end{align}
where the future state is predicted by the simulation function $f_s(s_t, a)$. Prior work uses supervised learning to approximate the actual value function with the trajectory data. In power grid management, we found an alternative estimate function that does not rely on approximation. The idea comes from that the unsolved overloaded power line can induce more overloaded power lines and even lead to large-scale blackouts (i.e., large penalty). We thus define a risk function to monitor the overloaded power lines:

\begin{align}
R_{isk} = max\frac{I_i}{I_{max_i}}, i=1,2,...,|L|,
\end{align}
where $I_i$ and $I_{max_i}$ represent the current flow and the flow capacity of line $L_i$, respectively. The ratio $\frac{I_i}{I_{max_i}} > 1$ means that the power line $i$ is overloaded. We replace the value function in Equation~\ref{eq:action_selection_0} with the risk function and have:
\begin{align}
a_t = \argmin_{a \in A}(R_{isk}(f_s(s_t,a)),
\label{eq:action_selection}
\end{align}

\subsection{Policy Optimization}
As shown in Equation~\ref{eq:action_selection}, action selection relies on the simulation function $f_s(s,a)$. If the $f_s(s,a)$ is known and differentiable, we can compute the backpropagating gradients and optimize the policy directly by maximizing the average reward in Equation~\ref{eq:rl_object}. However, it is often difficult to acquire the exact dynamics function $f_s(s,a)$ and it is unnecessarily differentiable in real-world applications. Though previous work uses an differentiable linear or nonlinear approximation function as an alternative \cite{jin2009model,larsson2002emergency}, it introduces additional noise into optimization, and the performance highly relies on the accuracy of the approximation function \cite{huang2019adaptive}. \\
To address the problem of obtaining the actual $f_s(s,a)$, we apply the black-box optimization of evolution strategies (ES) \cite{eigen1973ingo,schwefel1977numerische,salimans2017evolution} to update the policy, which does not require backpropagating gradients for optimization. ES is an optimization algorithm inspired by natural evolution: A population of parameter vectors is derived from current parameters and evaluated in the environment, after which the parameter vectors with the highest scores will be recombined and form the next generation. In SAS, we repeatedly inject Gaussian noise $\epsilon$ into the parameter vector of the original policy and obtain a bunch of policies for exploration. \\
Overall, The optimization process repeatedly runs the two following phrases until the policy converges: (1) obtain the exploratory policy by perturbing the policy parameters with $\theta + \epsilon$ and evaluating its performance in power grid management (2) collect the sampled noise parameters $\epsilon$ and the related rewards for the computation of combined gradient, and update the policy parameter.

\subsection{Discussion on Action Set Size}
We now discuss the selection of action set size $K \in [1, N]$, where N is the number of actions. We first discuss the boundary values. If the size K is equal to 1, the algorithm reduces to the traditional RL method, since the function $f_s(s_t,a)$ can be omitted in Equation~\ref{eq:action_selection} and we can perform backpropagation optimization to maximize the average return. When K is equal to N, the algorithm looks like an inefficient SPS system. It tries all the possible actions and selects the one with the lowest risk level, which is unacceptably time-consuming in the real-world and ignores the long-term return. The policy can also not be improved as the action selection always tests the whole action space, and policy is of no use. \\

Intuitively, a larger action set allows the algorithm to search more times at each step, which can improve the search result. We will further discuss the set size in the experiment section.
\begin{algorithm}
  \caption{Action Set based Optimization} \label{alg:set_based_optimization}
  \begin{algorithmic}[1]
   \REQUIRE Initialize the policy : $\pi_\theta$ \\
    Input learning rate $\alpha$, noise stand deviation $\sigma$, and \\ action set size $K$.
\REPEAT
    \FOR{i in $\{1,2...,n\}$}
    \STATE Sample Gaussian noise vector: $\epsilon \in N (0, I)$
    \STATE Perturb the policy with $ \epsilon_i * \sigma$: $\pi_{\theta + \epsilon_i * \sigma} (s, a)$
    \WHILE{not the end of the episode}
    \STATE Top-K actions with higher probabilities forms a set
    \STATE Select the action $a$ according to Equation~(\ref{eq:action_selection}) 
    \ENDWHILE
    \STATE Compute the total return $r_i$ \STATE Record the exploration result $(\epsilon_i, r_i)$
    \ENDFOR
    \STATE Summarize the gradient $g = \frac{1}{n\sigma}\sum_{i=1}^n r_i \sigma_i$
    \STATE Update the policy $\theta \leftarrow \theta + \alpha g$
\UNTIL{ $\pi_\theta$ converges}
\end{algorithmic}
\end{algorithm}
\subsection{Algorithm Summary}
Traditional search-based planning algorithms select the action mainly based on the reward estimation and exploration bonus, as shown in Equation~\ref{eq:planning_act_selection}. In order to consider the environmental constraints, we split the action selection into two steps. The policy first outputs a number of action candidates as the limited search space, and then the algorithm starts to search with the set and filters actions that violate the constraints. The policy will not get a positive feedback if all the action candidates fail to meet the constraints. We further introduce ES with black-box optimization to maximize the average return such that the policy learns to output actions with a higher return while strictly meeting the constraints. The SAS algorithm is summarized at Algorithm~\ref{alg:set_based_optimization}.

\section {Experiments}
To evaluate our algorithm, we compare it with the baseline reinforcement learning algorithms in the Grid2Op environment, including DQN \cite{mnih2015human}, APE-X \cite{horgan2018distributed}, and Semi-Markov
Afterstate Actor-Critic (SMAAC) \cite{yoon2021winning} recent proposed for power grid management. SMAAC tackles the challenge of large action and state space by introducing the afterstate representation \cite{sutton2018reinforcement}, which models the state after the agent has made the decision but before the environment has responded.

\begin{figure*}[t]
    \centering
    \includegraphics[width=\linewidth]{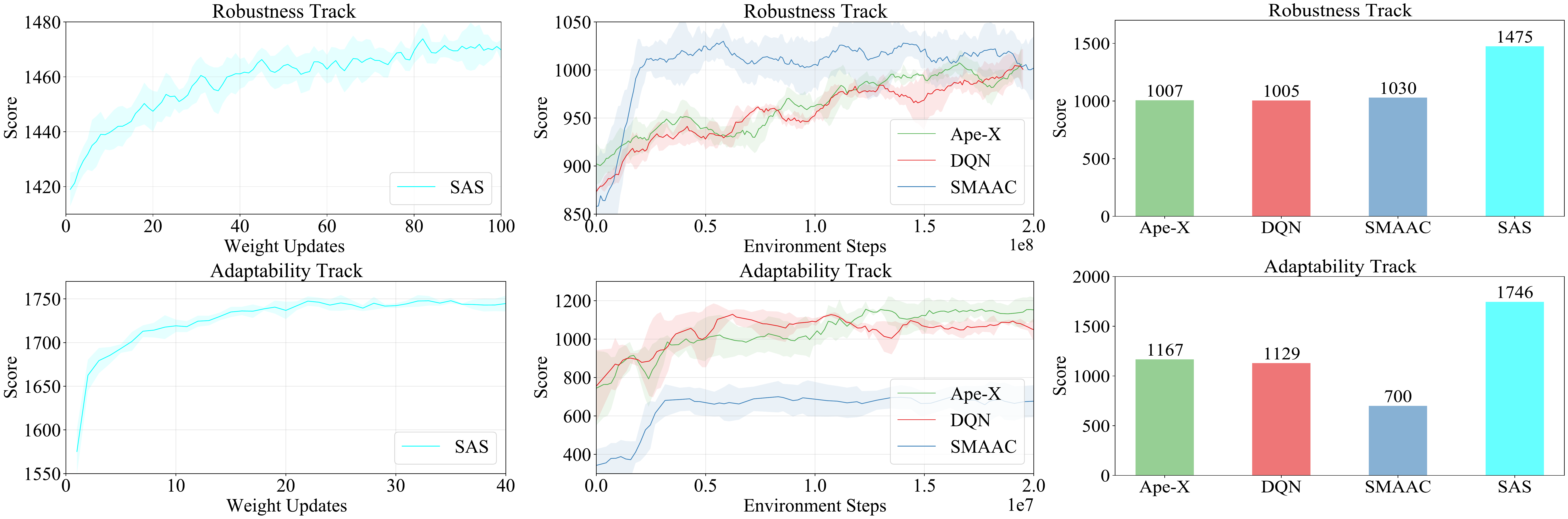}
    \caption{Evaluation of SAS and baseline methods on robustness and adaptability tracks. (Left) Average return of SAS over weight update times. (Mid) Average return of RL methods over environment steps. (Right) Performance comparison at convergence. We plot the figure of SAS separately as its policy updates at a much slower frequency than RL methods, which update every a few environment steps.}
    \label{fig:experiment}
\end{figure*}

\subsection{Experiment Setup}
The experiment includes two tasks in NeurIPS 2020 L2RPN: robustness and adaptability tracks. The power grid in the robustness task has 36 substations, 59 power lines, 22 generators and 37 loads, providing the grid state of 1266 dimensions and 66918 possible topological actions. The most distinguishing feature in this task is the unexpected attack on power lines. Some of the power lines will be disconnected suddenly every day at different times, which will collapse the grid if the agent cannot overcome the attack in a short time. \\
The adaptability task has a larger power grid, approximately three times that of the robustness track, with 118 substations, 186 lines, 62 generators and 99 loads. The task reflects the emerging deployment of renewable energy generators, and it evaluates the agent with the environments containing different amounts of renewable energy generators. The agent has to adapt to the increasing proportion of renewable energy. Note that the control flexibility decreases as the number of less controllable renewable energy generators increases.

\subsection{Implementation}
The policy network contains four fully connected layers with RELU as the activation function and outputs the probabilities of each action given the current grid state. We use the same network structure for all the baseline algorithms for a fair comparison. The policy produces 100 action candidates for action selection at the planning stage (i.e., K=100).\\
Following the parallel training implementation in ES \cite{salimans2017evolution}, our implementation employs a number of CPUs to finish the exploration and search processes in parallel. At each iteration, we generate a large number of exploration policies and distribute them into different machines. The machines evaluate the performance of the noisy policies, compute the accumulative rewards and return them to the learner. Then we collect the feedback from the machines and compute the combined gradient for policy optimization. We use 500 CPUs and 1 GPU for the distributed version.
Figure~\ref{fig:experiment} shows the performance of various algorithms throughout training in robustness and adaptability tasks. The shaded area shows one standard deviation of scores. Each experiment was run four times with different random seeds.

\begin{figure}
    \centering
    \subfigure[]{\includegraphics[width=0.495\textwidth]{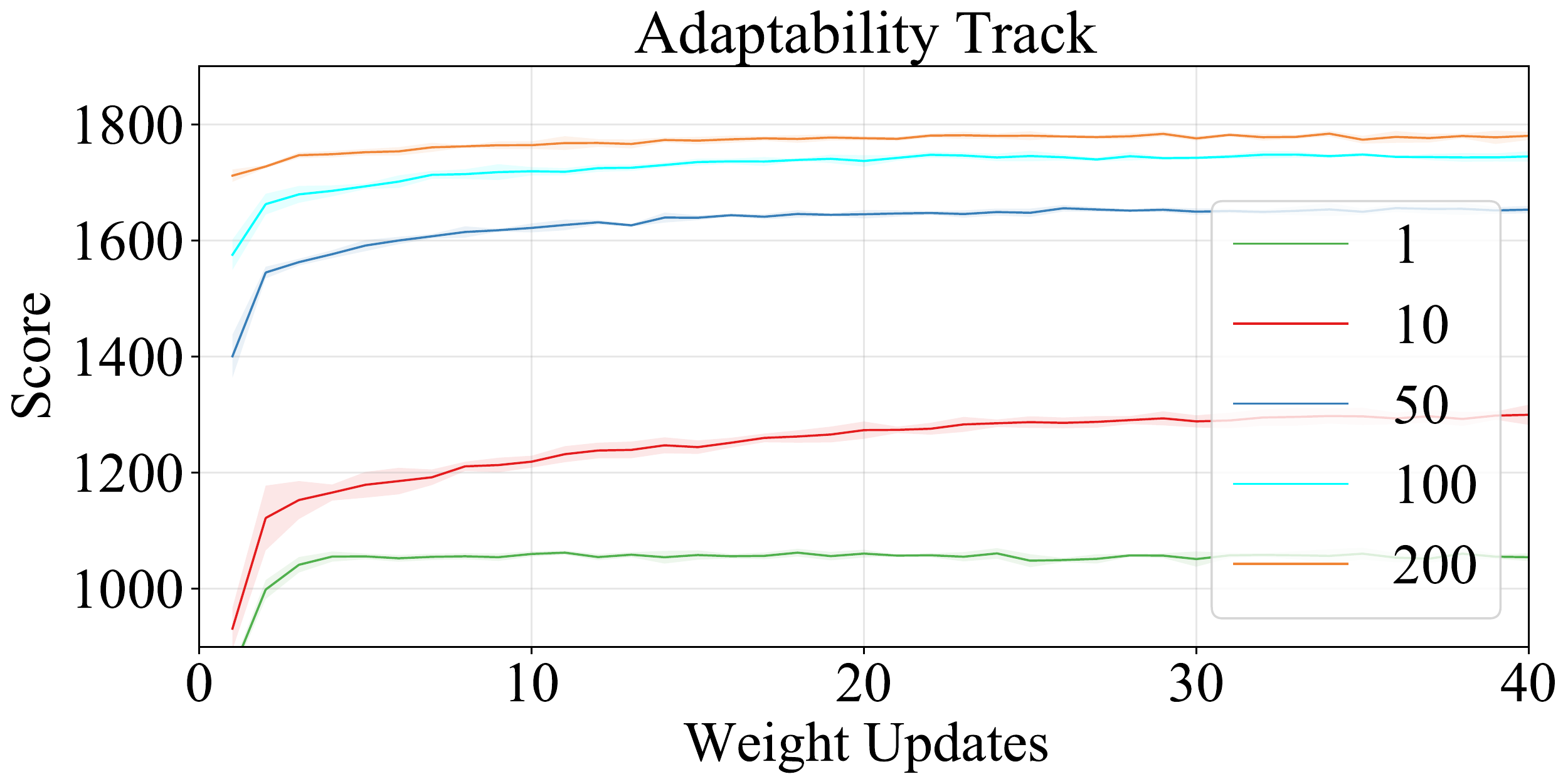}} 
    \subfigure[]{\includegraphics[width=0.495\textwidth]{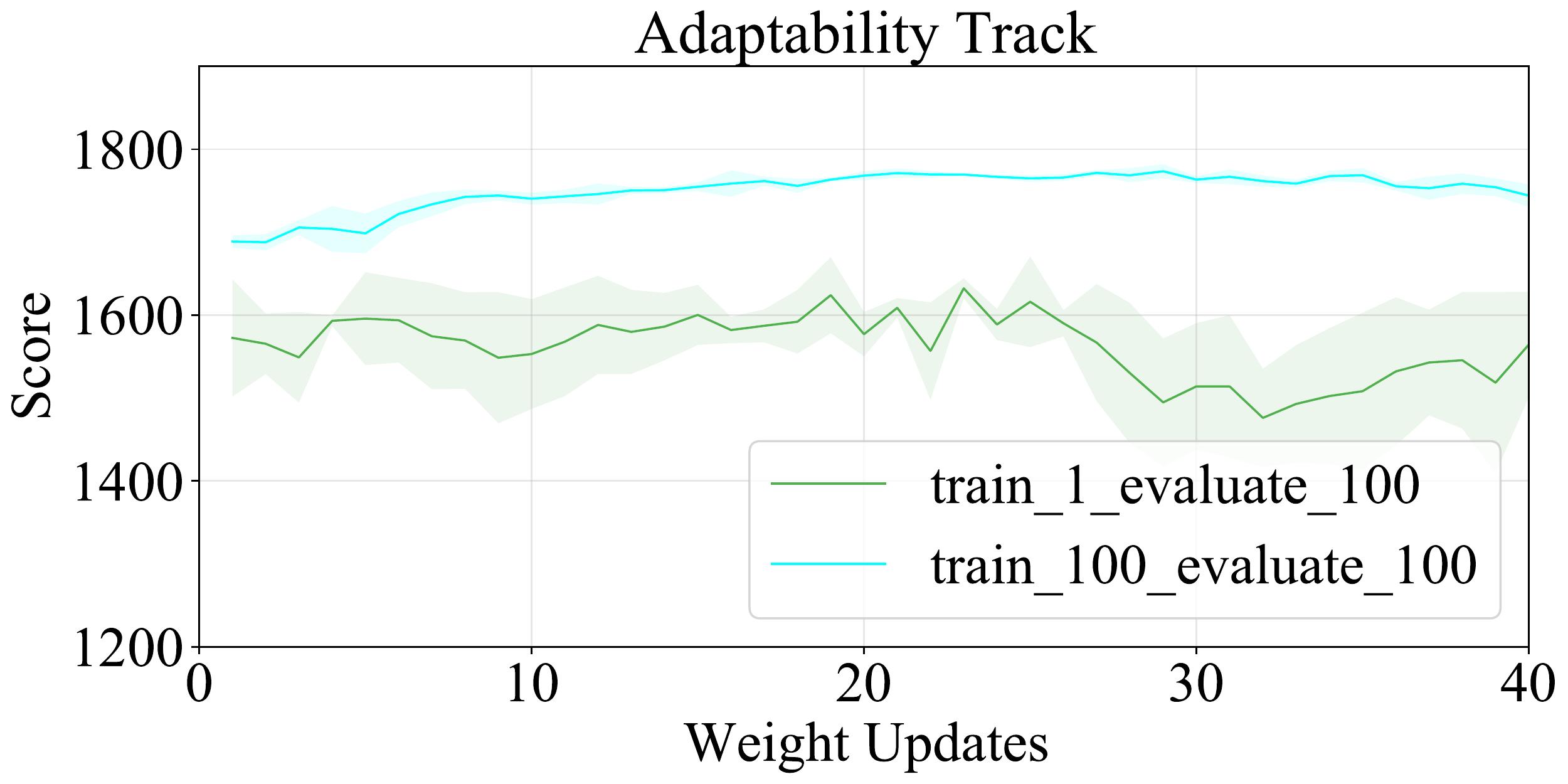}} 
    \vspace{-.2cm}
    \caption{(a) Training curves with different action set size (b) Re-evaluation of the learned policy with larger action set size.}
    \label{fig:ablation}
\end{figure}

SMAAC learns faster than DQN and APE-x in both tasks, as the after-state representation provides a sufficient representation of the grid than the traditional state-action pair representation \cite{yoon2021winning}. However, its performance is worse than other methods at the adaptability task. The possible explanation is that the distribution shift of renewable energy production makes it more difficult to predict the future return. Though SMAAC provides better representation for the state and action pair, it cannot help model the distribution change that cannot be observed through the state.
The SAS algorithm significantly outperforms the prior RL methods. Note that though SAS can achieve excellent performance in about one hundred iterations, a large amount of data (10 000 episodes) is required at each iteration. Since ES supports parallel training, we address the problem by using 500 CPUs for training, and it takes only about 1 hour for each iteration. \\
To better understand SAS, we measure how SAS performs with respect to various action set sizes and present the result in Figure~\ref{fig:ablation} (a). While the set size is equal to 1, the performance of SAS is similar to the previous RL method. As SAS searches with a larger action set, its performance rises steadily. Does that mean our performance gain comes from more search times? We further re-evaluate the learned policy with a larger action set size. As shown in Figure~\ref{fig:ablation} (b), though
the evaluation uses the same action set size (K=100), the policy performs better while it is trained with a larger action set. The empirical result shows that the policy learns to produce high-quality action candidates, which can improve the search efficiency (i.e., higher return in the same search time).

\begin{table}
\caption{Top 5 teams in NeurIPS2020 L2RPN  competition.}
\label{table:competition}

\begin{tabular*}{\textwidth}{l@{\extracolsep{\fill}}lllll}
Team               & \textbf{1(Ours)} & 2     & 3     & 4     & 5     \\ \hline
Robustness Track   & \textbf{59.26}  & 46.89 & 44.62 & 43.16 & 41.38 \\ \hline
Adaptability Track & \textbf{25.53}  & 24.66 & 24.63 & 21.35 & 14.01 \\ \hline
\end{tabular*}
\end{table}

\subsection{Competition}
There are two tracks in NeurIPS2020 L2RPN competition: robustness and adaptability tracks. We attended both tracks. Each submitted agent is tested in 24 unseen scenarios that cover every month of the year. The reward in each environment is re-scaled, and the total reward of 24 environments is used for ranking. 
As shown in Table~\ref{table:competition}, the SAS agent ranked first in both tracks.

\section {Conclusion}
In this paper, we propose a novel algorithm for grid management that searches within the action set decided by a policy network. By exploiting the simulation function to guarantee that selected actions strictly meet the constraints, the policy learn to adapts to the constraints while maximizing the reward in long run. To optimize the policy, we employed evolution strategies. With the proposed SAS algorithm, our agent outperformed prior RL approaches and won both tracks in the NeurIPS2020 L2RPN competition. Our work provides a novel approach to combine the complex environment constraints with policy optimization, which can potentially be applied to other real-world scenarios such as industry control and traffic control.

\clearpage

\bibliographystyle{splncs04}
\bibliography{sample-base}
\clearpage
\appendix
\section{Grid2Op Environment}
\label{sec:env}
Grid2Op \cite{marot2020l2rpn} is an open-source environment developed for testing the performance of controllers in power grid management. It simulates the physical power grid and follows the real-world power system operational constraints and distributions.\\
The environment provides interactive interfaces based on the gym library \cite{brockman2016openai}. At each episode, it simulates a period of time (e.g., a week or month) at the time interval of 5 minutes. At each time step, the controller receives the state of the power grid and takes actions to operate the grid if necessary. The simulation terminates at the end of the period or terminates prematurely if the controller fails to operate the grid properly, which can occur under two conditions: (1) Some actions split the grid into several isolated sub-grids.
(2) The electricity power transmitted from the stations cannot meet the consumption requirement of some loads.
Too many disconnected lines will significantly increase the risk of causing these two conditions. A power line gets disconnected automatically if the current flow exceeds the maximum limit for 3 time steps (i.e., 15 minutes). In this case, the power line cannot be reconnected until the end of the recovery time of 12 time steps. \\
Grid2Op has a large state space and action space. In addition to the exponentially increasing possible grid topology we mentioned, the grid state contains other topology features such as the current, voltage magnitude of each power line, generation power of each power station, the required power of each load. Though only one substation can be reconfigured at each time step (to simulate that a human or an expert can perform a limited number of actions in a time period), the number of available actions for topology reconfiguration is ${\sum_{i=1}^{i=M}2^{|S_i|}}$. In the NeurIPS2020 L2RPN competition, there are 118 substations with 186 power lines, which introduces over 70,000 discrete actions related to unique topology.\\
The reward setting in Grid2Op mainly relies on the reliability of the power grid. At each time step, the environment gives a bonus for safe management, and the controller will no longer gain positive rewards if it fails to manage the power network properly, which can lead to early termination of the episode. There are also costs (penalty) of operations. To encourage the controller to explore the operation flexibility on topology reconfiguration, the cost of topology change is much smaller than re-dispatching the power generation of the power plant.\\
In the physical world, the operators often use a simulation system to compute the possible outcomes of actions to control risk \cite{bernard1996735,trudel1999hydro,marot2020l2rpn}, and Grid2Op also provides a similar function named \textit{simulate}, which can mimic the one-step operational process. It allows the user to check if the action violates the power network constraint (e.g., if the target power generation exceeds the maximum output of the power plant).  Note that this function can only be called once at each time step (i.e., one-step simulation), and its prediction on future states may bias from the actual state. 

\end{document}